\documentclass[a4paper,USenglish,cleveref,autoref,thm-restate]{lipics-v2021}

\DeclareUnicodeCharacter{00A7}{\S}
\DeclareUnicodeCharacter{00B1}{\ensuremath{\pm}}
\DeclareUnicodeCharacter{00B9}{\textsuperscript{1}}
\DeclareUnicodeCharacter{00D7}{\ensuremath{\times}}
\DeclareUnicodeCharacter{0394}{\ensuremath{\Delta}}
\DeclareUnicodeCharacter{03BA}{\ensuremath{\kappa}}
\DeclareUnicodeCharacter{2013}{--}
\DeclareUnicodeCharacter{2014}{---}
\DeclareUnicodeCharacter{201C}{``}
\DeclareUnicodeCharacter{201D}{''}
\DeclareUnicodeCharacter{2192}{\ensuremath{\to}}
\DeclareUnicodeCharacter{2212}{\ensuremath{-}}
\DeclareUnicodeCharacter{2265}{\ensuremath{\ge}}

\usepackage{array}
\usepackage{booktabs}
\usepackage{longtable}
\usepackage{pdflscape}
\usepackage{tikz}
\usepackage{pgfplots}
\usetikzlibrary{arrows.meta,calc,fit,positioning,shapes.geometric}
\pgfplotsset{compat=1.18}

\newcolumntype{L}[1]{>{\raggedright\arraybackslash}p{#1}}

\newcommand{\systemname}{SCM}
\newcommand{\comparatorname}{Production Comparator}
\definecolor{supraRed}{rgb}{0.8667,0.0784,0.2196}
\definecolor{supraInk}{rgb}{0.0902,0.1059,0.1412}
\definecolor{supraMuted}{rgb}{0.4039,0.4431,0.4863}
\definecolor{supraLine}{rgb}{0.6431,0.6863,0.7255}
\definecolor{supraPanel}{rgb}{0.9569,0.9647,0.9686}
\definecolor{supraGreen}{rgb}{0.5176,0.7882,0.2196}
\definecolor{supraBlue}{rgb}{0.2471,0.6275,0.9569}
\definecolor{supraOrange}{rgb}{1.0000,0.6745,0.2196}
\definecolor{supraPurple}{rgb}{0.4235,0.2706,0.9176}
\tikzset{
  scmblock/.style={draw=supraInk!80, rounded corners=2pt, align=center, inner xsep=5pt, inner ysep=4pt, font=\sffamily\scriptsize\bfseries, fill=white, text=supraInk, line width=0.45pt},
  scmproc/.style={scmblock, draw=supraRed!85!black, fill=supraRed, text=white, minimum width=2.1cm},
  scmdark/.style={scmblock, draw=supraInk, fill=supraInk, text=white},
  scmstore/.style={scmblock, draw=supraLine, fill=supraPanel, text=supraInk},
  scmghost/.style={scmblock, draw=supraLine, fill=white, text=supraMuted, dashed},
  scmnote/.style={font=\sffamily\scriptsize\bfseries, align=center, text=supraInk},
  scmarrow/.style={-{Latex[length=2.2mm,width=1.35mm]}, line width=0.45pt, draw=supraLine},
  scmaccentarrow/.style={scmarrow, draw=supraRed},
  scmdashedarrow/.style={scmarrow, dashed, draw=supraRed!80}
}
\makeatletter
\nolinenumbers
\hideLIPIcs
\AtBeginDocument{\definecolor[named]{lipicsYellow}{rgb}{0.8667,0.0784,0.2196}}

\DeclareCaptionLabelFormat{boxed}{%
  \kern0.05em{\color{supraRed}\rule{0.73em}{0.73em}}%
  \hspace*{0.67em}\bothIfFirst{#1}{~}#2}
\newcommand{\supraWhitepaperAuthors}{Joshua Tobkin\textsuperscript{*} and David Yang}
\newcommand{\supraWhitepaperCorrespondingAuthor}{%
  \textsuperscript{*}Corresponding author:
  \href{mailto:j.tobkin@supra.com}{j.tobkin@supra.com}}
\newcommand{\supraWhitepaperAffiliation}{Supra Research}
\newcommand{\supraWhitepaperDate}{17 July 2026}

\def\supra@footer{%
  \hfil\raisebox{-1.5mm}{\includegraphics[height=10mm]{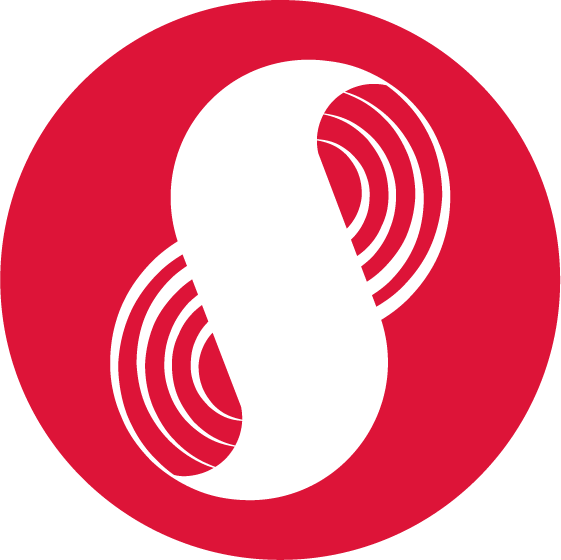}}\hfil}

\def\@sect#1#2#3#4#5#6[#7]#8{%
  \ifnum #2>\c@secnumdepth
    \let\@svsec\@empty
  \else
    \refstepcounter{#1}%
    \protected@edef\@svsec{\@seccntformat{#1}\relax}%
  \fi
  \@tempskipa #5\relax
  \ifdim \@tempskipa>\z@
    \begingroup
      #6{%
        \@hangfrom{\hskip #3\relax
          \ifnum #2=1
            \colorbox{lipicsYellow}{\kern0.15em\textcolor{white}{\@svsec}\kern0.15em}\quad
          \else
            \@svsec\quad
          \fi}%
          \interlinepenalty \@M #8\@@par}%
    \endgroup
    \csname #1mark\endcsname{#7}%
    \addcontentsline{toc}{#1}{%
      \ifnum #2>\c@secnumdepth \else
        \protect\numberline{\csname the#1\endcsname}%
      \fi
      #7}%
  \else
    \def\@svsechd{%
      #6{\hskip #3\relax
      \@svsec #8}%
      \csname #1mark\endcsname{#7}%
      \addcontentsline{toc}{#1}{%
        \ifnum #2>\c@secnumdepth \else
          \protect\numberline{\csname the#1\endcsname}%
        \fi
        #7}}%
  \fi
  \@xsect{#5}}

\def\ps@plain{%
  \let\@mkboth\@gobbletwo
  \let\@oddhead\@empty
  \let\@evenhead\@empty
  \let\@oddfoot\@empty
  \let\@evenfoot\@empty}

\def\ps@suprawhitepaper{%
  \def\@evenhead{\large\sffamily\bfseries
                 \llap{\hbox to0.5\oddsidemargin{\thepage\hss}}\leftmark\hfil}%
  \def\@oddhead{\large\sffamily\bfseries\rightmark\hfil
                \rlap{\hbox to0.5\oddsidemargin{\hss\thepage}}}%
  \let\@oddfoot\supra@footer
  \let\@evenfoot\supra@footer
  \let\@mkboth\markboth
  \let\sectionmark\@gobble
  \let\subsectionmark\@gobble}
\def\@maketitle{%
  \newpage
  \null\vskip-\baselineskip
  \vskip-\headsep
  \@titlerunning
  \@authorrunning
  \parindent\z@ \raggedright
  {\LARGE\sffamily\bfseries\mathversion{bold}\@title
   \if!\@subtitle!\else{\\\Large\sffamily\bfseries\mathversion{bold}\@subtitle}\fi \par}%
  \vskip 1em
  {\Large\bfseries\supraWhitepaperAuthors\par}%
  \vskip 0.25em
  {\normalsize\supraWhitepaperCorrespondingAuthor\par}%
  \vskip 0.25em
  {\large\supraWhitepaperAffiliation\par}%
  \vskip 0.95em
  {\large\supraWhitepaperDate\par}%
  \par}

\renewenvironment{abstract}{%
  \vskip\topmattervskip\bigskipamount
  \noindent
  \rlap{\color{lipicsLineGray}\vrule\@width\textwidth\@height1\p@}%
  \hspace*{7mm}\fboxsep1.5mm\colorbox[rgb]{1,1,1}{\raisebox{-0.4ex}{%
    \large\selectfont\sffamily\bfseries\abstractname}}%
  \vskip3\p@
  \fontsize{9}{12}\selectfont
  \setlength{\parindent}{1.5em}\indent\ignorespaces}
  {\vskip\topmattervskip\baselineskip\noindent}
\makeatother

\title{Supra Cognitive Modes: A Routed Architecture for Agent Memory}
\titlerunning{Supra Cognitive Modes}

\author{Joshua Tobkin}{Supra Research}{j.tobkin@supra.com}{}{}
\author{David Yang}{Supra Research}{}{}{}
\authorrunning{J. Tobkin and D. Yang}
\Copyright{Supra Research}


\category{}
\relatedversion{}

\EventEditors{}
\EventNoEds{0}
\EventLongTitle{}
\EventShortTitle{}
\EventAcronym{}
\EventYear{2026}
\EventDate{}
\EventLocation{}
\EventLogo{}
\SeriesVolume{}
\ArticleNo{}

\AtBeginDocument{%
  \hypersetup{%
    pdfsubject={Branded preprint on a routed architecture for agent memory},%
    pdfcreator={LaTeX},%
    pdfproducer={pdfTeX}%
  }%
}

\begin{document}
\maketitle

\begin{abstract}
Agent-memory workloads mix direct factual lookup, relation-chain and current-state reasoning, and broad synthesis over long histories. We describe Supra Cognitive Modes (\systemname), an architecture that maps explicit or automatically selected per-query modes to retrieval and synthesis payloads over one shared ingest substrate. A frozen semantic classifier and runtime gates dispatch queries among fused lexical and dense lookup, graph or iterative multi-hop handling, and stratified long-form synthesis. The substrate combines multi-granularity embeddings, extracted triples, fact-version metadata, and optional asynchronous enrichments.

We characterize the deployed configuration on three benchmarks: Long-term Conversational Memory (LoCoMo; $n=1{,}986$), MemoryAgentBench (MAB; $n=3{,}671$), and LongMemEval ($n=500$). The reference run records 84.87\% on LoCoMo factoid categories and 68.61\% on adversarial abstention, 61.49\% on MAB across two repetitions, and 86.00\% on LongMemEval. A repository-backed reproduction produces similar aggregate scores and supports task- and mode-conditioned failure analysis. Raw baseline outputs, aligned end-to-end timing for LoCoMo and LongMemEval, and complete token ledgers are unavailable; stored rows also omit some final runtime decisions. The results characterize one implemented routed configuration and its diagnostic failure patterns, while source inspection verifies the per-query control interface and shared-substrate design. Causal routing effects, efficiency gains, and statistical significance remain outside the available evidence.
\end{abstract}
\section{Introduction}

Language-model agents accumulate conversations, retrieved evidence, and user-specific state over long periods. Their memory layer must answer heterogeneous questions: a direct lookup may need one recent fact, a relation-chain question may require several linked records, and a broad request may require synthesis across hundreds of conversations. A single retrieval and synthesis policy is unlikely to serve all three shapes equally well.

This heterogeneity creates a deployment tension among answer quality, response time, and cost. Aggressive ingest-time extraction can reduce query-time work but pays before the future query volume is known. Broad multipass synthesis can improve coverage but adds latency and model usage to questions that may need only one fact. We use this memory trilemma as a design lens, not as a claim that the evaluated configurations establish a universal accuracy--latency--cost frontier.

\subsection{Approach}

Supra Cognitive Modes (\systemname) treats the memory operating point as a per-query control. An application can pass an explicit mode or request automatic selection. A \emph{mode} is a semantic label, a \emph{payload} is the concrete retrieval and synthesis configuration derived from that label, and a \emph{procedure} is the execution family that answers the query. Modes and procedures are not one-to-one: several labels can map to the same initial payload, and runtime gates can change the executed procedure.

The evaluated configuration uses four frozen semantic labels for single-fact lookup, long-form synthesis, time-anchored lookup, and latest-version resolution. A runtime route handles in-context-learning corpora, and query-shape gates can force multi-hop handling. These controls dispatch among direct fused retrieval, graph or iterative relation handling, and stratified long-form synthesis. All procedures read from one substrate containing multi-granularity embeddings, extracted relations, fact-version metadata, and optional cached enrichments.

Source and configuration inspection establish that this per-query control surface and the procedure families exist. Completed benchmark runs characterize one deployed configuration and expose diagnostic task and mode patterns. Because stored rows do not preserve every final runtime decision and fixed-policy controls were not run, the results do not identify routing or an individual procedure as the cause of a benchmark difference.

\subsection{Contributions}

This paper makes three bounded contributions:

\begin{enumerate}
\item It defines a mode--payload--procedure interface that exposes an agent-memory operating point per query rather than fixing one policy for an entire deployment.
\item It describes a source-visible implementation of that interface over a shared asynchronous substrate, including direct, graph-capable, and long-form procedure families.
\item It provides a repository-traceable characterization of the deployed configuration across three memory benchmarks, together with reproducible task- and mode-conditioned failure strata and an explicit audit of the available timing, cost, judge, and provenance evidence.
\end{enumerate}

The architectural novelty is the composition and control interface, not the underlying retrieval primitives. Hybrid lexical and dense retrieval, reciprocal-rank fusion, graph traversal, temporal metadata, and multipass synthesis come from established lines of work discussed next.

\section{Related Work}

\subsection{Agent memory systems}

Foundational agent architectures established that stored experience can shape planning and behavior. Generative Agents stores observations and synthesized reflections for later planning \cite{park2023generativeagents}; CoALA organizes language agents around modular memory, actions, and decision processes \cite{sumers2023coala}; Reflexion reuses verbal self-critique \cite{shinn2023reflexion}; and Voyager accumulates a reusable skill library \cite{wang2023voyager}. MemoryBank adds long-term conversational memory, updating, and forgetting \cite{zhong2024memorybank}, while MemGPT and Letta make memory an explicit resource across main-context, recall, and archival tiers \cite{packer2023memgpt,letta2026memorydocs}.

Recent systems place more work in memory construction and organization. Mem0 extracts persistent user-specific memories at ingest \cite{chhikara2025mem0}; Zep/Graphiti uses a temporal knowledge graph with provenance \cite{rasmussen2025zep}; A-Mem organizes memories as dynamically linked notes \cite{xu2025amem}; and MIRIX uses several typed stores under a controller \cite{wang2025mirix}. MemOS frames memory as an operating-system resource \cite{li2025memos}, while LightMem moves consolidation into lightweight and offline stages \cite{fang2026lightmem}. These systems show that memory quality depends not only on retrieval but also on where construction, organization, and synthesis costs are paid.

\subsection{Adaptive retrieval and routing}

Routing among computation paths is well established. Adaptive-RAG classifies questions into no-retrieval, single-step, and multi-step tiers \cite{jeong2024adaptiverag}; Self-RAG learns when to retrieve and critique evidence \cite{asai2023selfrag}; IRCoT interleaves retrieval with multi-hop reasoning \cite{trivedi2023ircot}; and FLARE retrieves when generation becomes uncertain \cite{jiang2023flare}. FrugalGPT and RouteLLM apply a parallel idea to model and cascade selection under cost--quality trade-offs \cite{chen2023frugalgpt,ong2024routellm}.

\systemname{} differs primarily in the routed unit. Rather than exposing only retrieval depth or model choice, it maps a per-query semantic label to a payload that can select a retrieval strategy, substrate reads, prompt family, and synthesis procedure. This paper characterizes the deployed routed configuration but does not compare it with fixed-policy controls.

\subsection{Retrieval, graph, and synthesis primitives}

The procedure families reuse established components. Retrieval-augmented generation and dense passage retrieval provide the dense-retrieval foundation \cite{lewis2020rag,karpukhin2020dpr}; sparse lexical retrieval and reciprocal-rank fusion combine complementary rankings \cite{cormack2009rrf}. GraphRAG and RAPTOR demonstrate graph- and hierarchy-based synthesis over large text collections \cite{edge2024graphrag,sarthi2024raptor}. Temporal knowledge graphs preserve changing relations and provenance \cite{cai2023tkgcsurvey,rasmussen2025zep}. The contribution here is not a new primitive, but the organization of these mechanisms into procedure-owned paths over one shared substrate, including latest-revision precedence for ordinary current-state queries.

\subsection{Benchmarks and judge context}

The evaluation uses three complementary suites. Long-term Conversational Memory (LoCoMo) studies very long conversations \cite{maharana2024locomo}; LongMemEval covers extraction, multi-session reasoning, temporal reasoning, knowledge updates, and abstention \cite{wu2024longmemeval}; and MemoryAgentBench (MAB) covers retrieval, test-time learning, long-range understanding, and selective forgetting \cite{hu2025memoryagentbench}. Newer suites broaden the surface toward factual and reflective memory, million-token conversations, web-agent experience, and long-horizon trajectories \cite{tan2025membench,tavakoli2025beam,wu2026longmemevalv2,zhao2026amabench}.

Several benchmark metrics use language-model judges. Prior judge evaluations document prompt sensitivity, position and verbosity effects, and imperfect human agreement \cite{zheng2023judge}. We preserve the benchmark prompt families and model versions, separate lexical and judge-based cuts, and interpret judge results as benchmark-metric agreement rather than independently validated correctness.

\subsection{Positioning}

The closest prior systems already contain memory tools, graphs, consolidation policies, or internal routers. The narrower distinction studied here is an application-visible per-query selector whose procedures share one ingest substrate. The evaluation contribution is correspondingly bounded: it reports descriptive accuracy and diagnostic traces while separating those observations from unavailable fixed-policy, aligned timing, complete cost, and mechanism-level evidence.

\section{System Design}

\subsection{Design objective}

The architecture treats accuracy, query latency, and cost placement as coupled design dimensions. A direct lookup should not require the same evidence pool and synthesis plan as a cross-session report, while a current-state query may need revision handling that an ordinary lookup does not. \systemname{} therefore exposes the operating point as a per-query selector over one shared memory substrate.

The design separates four concepts. The \emph{mode source} is either an application-selected label or the benchmark classifier. A \emph{payload map} turns that label into concrete retrieval and synthesis flags. \emph{Runtime gates} can revise the initial tier using query or retrieved-context signals. Finally, a \emph{procedure} executes direct lookup, graph or iterative multi-hop handling, or long-form synthesis. This separation matters because semantic modes and executed procedures are not one-to-one.

\begin{figure}[htbp]
\centering
\resizebox{\textwidth}{!}{%
\begin{tikzpicture}[node distance=8mm and 10mm, line cap=round, line join=round]
\node[scmdark, text width=1.65cm] (query) {Query};
\node[scmblock, text width=2.25cm, right=of query] (selector) {Mode source\\explicit or auto};
\node[scmblock, text width=2.35cm, right=10mm of selector] (classifier) {Payload map\\+ runtime gates};
\node[scmproc, text width=2.05cm, right=13mm of classifier, yshift=12mm] (simple) {Direct\\lookup};
\node[scmproc, text width=2.05cm, right=13mm of classifier] (multi) {Graph / iterative\\multi-hop};
\node[scmproc, text width=2.05cm, right=13mm of classifier, yshift=-12mm] (summary) {Long-form\\synthesis};
\node[scmstore, text width=3.75cm, below=17mm of multi] (substrate) {Shared substrate\\embeddings, triples, supplements};
\node[scmblock, text width=2.15cm, right=13mm of multi] (ground) {Synthesis /\\answer policy};
\node[scmdark, text width=1.75cm, right=of ground] (answer) {Answer\\or abstain};

\draw[scmarrow] (query) -- (selector);
\draw[scmarrow] (selector) -- (classifier);
\draw[scmaccentarrow] (classifier.east) -- (multi.west);
\draw[scmaccentarrow] (classifier.east) |- (simple.west);
\draw[scmaccentarrow] (classifier.east) |- (summary.west);
\draw[scmarrow] (simple.east) -- (ground.west);
\draw[scmarrow] (multi.east) -- (ground.west);
\draw[scmarrow] (summary.east) -- (ground.west);
\draw[scmarrow] (ground) -- (answer);
\draw[scmdashedarrow] (substrate.north) -- (multi.south);
\draw[scmdashedarrow] (substrate.north west) -- (simple.south);
\draw[scmdashedarrow] (substrate.north east) -- (summary.south);
\end{tikzpicture}%
}
\caption{Mode-to-procedure architecture. An explicit or automatic mode selects an initial payload; runtime gates may change the tier before a procedure reads from the shared substrate.}
\label{fig:selector-procedures}
\end{figure}
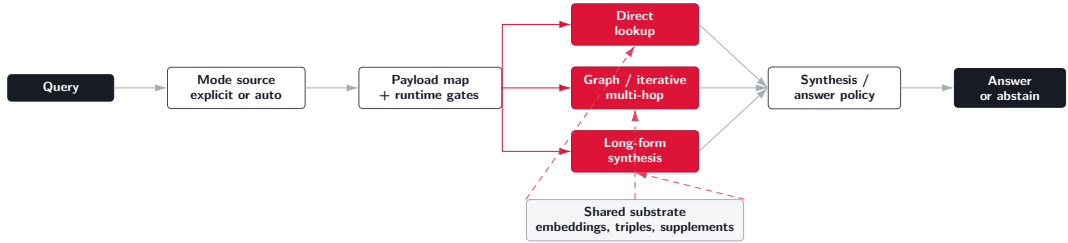

\subsection{Mode and runtime dispatch}

The frozen classifier emits one of four semantic labels and defaults to single-fact lookup when its output cannot be parsed. It was selected from five prompts on a synthetic off-test calibration set and then checked on a consensus-labeled development slice. Table~\ref{tab:compact-mode-map} shows that the semantic label set is larger than the initial payload action set: three labels share the same forced-simple configuration. Appendix C records the classifier evidence and complete mapping.

\begin{table}[htbp]
\small
\centering
\begin{tabular}{L{0.24\linewidth} L{0.23\linewidth} L{0.22\linewidth} L{0.21\linewidth}}
\toprule
Mode source & Retrieval config & Initial execution & Role in the reviewed run \\
\midrule
Single-fact, time-anchored, or latest-version label & \path{h40-temporal-synth} & Forced simple tier; top 100; Sonnet synthesis & Three semantic labels sharing one initial payload \\
Long-form label & \texttt{h17-narrative} & Graph-capable; top 100; Sonnet synthesis & Narrative retrieval and bullets-expand flags \\
Examples route & \path{h40-temporal-synth} & Forced simple tier; top 100 & Runtime \path{PATTERN_FROM_EXAMPLES} route for in-context-learning corpora \\
\bottomrule
\end{tabular}
\caption{Compact mapping from semantic modes to initial runtime payloads. Runtime gates and fallbacks can still change the final procedure.}
\label{tab:compact-mode-map}
\end{table}

The eval route can further use a lightweight tier classifier, intent rules, and a relation-chain detector. Multiple-choice questions can be forced to the simple tier, summary requests to the summary tier, and queries with several relation markers to multi-hop handling. Retrieval-shape detection can also reroute in-context-learning corpora after inspecting returned chunks. The dispatch logic is summarized as

\begin{verbatim}
mode = explicit_mode or classify(query)
payload = mode_to_payload(mode)
payload = runtime_gates(payload, query, retrieved_context)
procedure = procedure_for(payload.final_tier)
return execute(procedure, query, payload)
\end{verbatim}

The active long-form payload has a provenance mismatch: its JSON field records \texttt{graph\_routed}, while an amendment note records a change to \texttt{graph\_then\_iterative}. Stored outputs do not retain the final graph strategy. We therefore describe the supported procedure family without assigning an individual score to either strategy.

\subsection{Procedure families}

\noindent\textbf{Direct lookup.}
The direct path embeds the query and can add two lightweight paraphrases before running retrievers in parallel. The active configurations combine Okapi BM25 with dense retrieval over chunk, sentence, and paragraph indexes, then merge candidates with reciprocal-rank fusion. A deterministic paper-grade setting disables wall-clock recency boosting; the narrative configurations retain an entity-frequency reranker. The final depth is capped at 100 chunks. A single Sonnet call synthesizes the answer and receives an instruction to emit ``No information available'' when the queried entity is unsupported. That abstention behavior is prompt-directed rather than a deterministic post-retrieval filter.

\medskip
\noindent\textbf{Graph and iterative multi-hop handling.}
The multi-hop family targets relation chains and current-state questions. Extracted triples store normalized subject, predicate, and object fields together with source memory, confidence, version number, active status, and optional temporal fields. A planner can select a seed entity and hop sequence over a closed predicate vocabulary; a grounded walker then chooses among actual incoming and outgoing edges. Candidate triples are ordered by descending version number, giving newer values precedence for ordinary current-state retrieval while retaining older evidence for provenance. Depending on the active strategy, failed graph planning can fall back to iterative dense retrieval or standard fused retrieval.

\medskip
\noindent\textbf{Long-form synthesis.}
The summary tier samples the corpus chronologically, targeting 200 chunks, and applies a multipass bullets-expand procedure. A cached or online source-shape classifier distinguishes narrative, research, technical, log, and mixed corpora. The selected shape controls a coverage-planning prompt and an expansion prompt. The active payload enables shape-aware wrapping, broad research prompts, narrative name fidelity, and soft keypoint hints while leaving document-summary, canonical-entity, event-chain, and question-pattern reranking disabled.

\subsection{Shared asynchronous substrate}

All three procedure families read from one ingest substrate. The mandatory Tier-1 path writes chunks, optional sentence and paragraph segments, embeddings, lexical-index fields, and extracted triples. Once that path completes, fused lookup is available immediately; graph quality depends on the retained triples, and long-form synthesis can operate directly from stratified chunks.

Optional supplement jobs then cache higher-level signals such as document shape, anticipated keypoints, canonical entities, event chains, question patterns, structural importance, and entity types. These fields are nullable, and query code reads them only when the active payload enables the corresponding flag. Missing enrichments therefore degrade to online detection or omission rather than blocking query readiness.

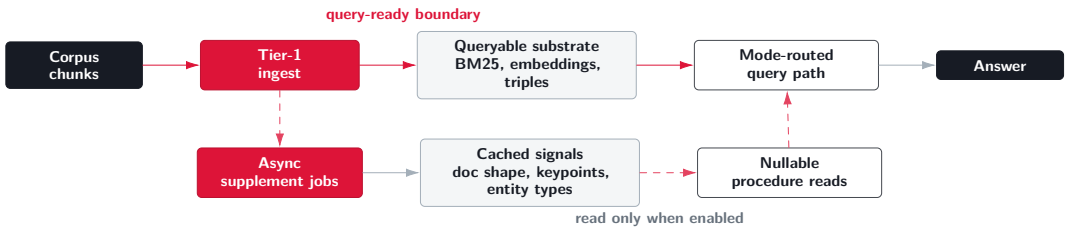
\begin{figure}[htbp]
\centering
\resizebox{\textwidth}{!}{%
\begin{tikzpicture}[node distance=7mm and 9mm, line cap=round, line join=round]
\node[scmdark, text width=1.8cm] (corpus) {Corpus\\chunks};
\node[scmproc, text width=2.15cm, right=of corpus] (tierone) {Tier-1\\ingest};
\node[scmstore, text width=3.1cm, right=of tierone] (substrate) {Queryable substrate\\BM25, embeddings,\\triples};
\node[scmblock, text width=2.55cm, right=of substrate] (query) {Mode-routed\\query path};
\node[scmdark, text width=1.65cm, right=of query] (answer) {Answer};
\node[scmproc, text width=2.25cm, below=9mm of tierone] (jobs) {Async\\supplement jobs};
\node[scmstore, text width=3.1cm, right=of jobs] (cache) {Cached signals\\doc shape, keypoints,\\entity types};
\node[scmblock, text width=2.55cm, right=of cache] (reads) {Nullable\\procedure reads};

\draw[scmaccentarrow] (corpus) -- (tierone);
\draw[scmaccentarrow] (tierone) -- (substrate);
\draw[scmaccentarrow] (substrate) -- (query);
\draw[scmarrow] (query) -- (answer);
\draw[scmdashedarrow] (tierone) -- (jobs);
\draw[scmarrow] (jobs) -- (cache);
\draw[scmdashedarrow] (cache) -- (reads);
\draw[scmdashedarrow] (reads) -- (query);
\node[scmnote, text=supraRed, anchor=south] at ($(tierone)!0.5!(substrate)+(0,0.55)$) {query-ready boundary};
\node[scmnote, text=supraMuted, anchor=north] at ($(cache)!0.5!(reads)+(0,-0.5)$) {read only when enabled};
\end{tikzpicture}%
}
\caption{Shared-substrate lifecycle. Mandatory ingest creates the queryable indexes; asynchronous supplements later add nullable signals read by selected payloads.}
\label{fig:substrate-lifecycle}
\end{figure}

The source establishes the presence of these procedures and reads, but the completed benchmark outputs do not isolate their individual effects. In particular, final runtime procedures and fallbacks are not fully persisted, and no component-removal controls are part of the retained evaluation.

\section{Evaluation Protocol and Evidence Boundaries}
\subsection{Evaluated answer paths}
The evaluation package supplies results for three configurations: \systemname{} in its deployed mode-routed configuration, a production comparator, and Mem0 v2 OSS with its language-model reranker. The repository contains \systemname{} outputs but not raw baseline outputs, so baseline parity cannot be independently verified. The table records the declared models and scoring settings used by the reproduction workflow. Answer generation is benchmark-specific: MAB scores the online route answer, whereas LoCoMo and LongMem score a later offline synthesis answer.

\begin{table}[htbp]
\small
\centering
\begin{tabular}{L{0.21\linewidth} L{0.23\linewidth} L{0.24\linewidth} L{0.22\linewidth}}
\toprule
Benchmark & Retrieved evidence & Scored answer & Timing and usage aligned? \\
\midrule
MemoryAgentBench & Online mode-routed eval path & Answer returned by the same request & Yes for \systemname{}; per-request wall time under concurrency \\
LoCoMo & Mode-routed retrieval snapshot & Offline Stage-3 synthesis & No; stored latency belongs to the earlier online answer \\
LongMemEval & Mode-routed retrieval snapshot & Offline Stage-3 synthesis & No; Stage-3 duration and usage were not persisted \\
\bottomrule
\end{tabular}
\caption{Benchmark-specific answer paths. Only the MAB output aligns the scored answer with the retained request timing.}
\label{tab:evaluated-answer-paths}
\end{table}

\subsection{Declared settings and non-invariants}

\begin{longtable}{L{0.32\linewidth} L{0.60\linewidth}}
\caption{Declared evaluation settings.}\label{tab:declared-evaluation-settings}\\
\toprule
Component & Value \\
\midrule
\endfirsthead
\toprule
Component & Value \\
\midrule
\endhead
Synthesis model & Claude Sonnet 4.5 (\texttt{claude-sonnet-4-5-20250929}, temperature 0) \\
\systemname{} synthesis & Online mode-aware answer generation for MAB; offline intent- and role-routed Stage-3 synthesis for LoCoMo and LongMem \\
Embeddings & OpenAI \texttt{text-embedding-3-large} at 1,024 dimensions \\
Memory-agent task suite scorer & Byte-faithful upstream substring-and-exact-match scorer \\
Memory-agent task suite synthesis judge & \texttt{gpt-4o-2024-05-13} at temperature 0.1, three-prompt keypoint-recall F1 \\
Long-conversational-memory factoid judge & \texttt{gpt-4o-mini-2024-07-18} with the comparator's accuracy prompt \\
Long-conversational-memory abstention scorer & Benchmark authors' paper-lexical substring scorer \\
Longitudinal personal-memory judge & \texttt{gpt-4o-2024-08-06} with the benchmark authors' six task-type prompts \\
Mem0 v2 SDK & \texttt{mem0ai==2.0.3}, \texttt{top\_k=100}, \texttt{rerank=True}, language-model reranker using \texttt{gpt-4o-mini} at temperature 0 \\
Mem0 v2 extraction model & Claude Haiku 4.5 \\
Vector store & ChromaDB (Mem0 v2); pgvector (\comparatorname{} and \systemname{}) \\
\bottomrule
\end{longtable}

The declared settings are intended to reduce several obvious confounds, but the absence of baseline rows prevents an audit of whether model, prompt, and generation parity held in the supplied baseline runs. Retrieval outputs, prompt inputs, vector stores, and pipeline structure can also interact with the generator and judges. We therefore report supplied configuration-level differences and avoid attributing them to individual memory components.

\subsection{Timing boundary}
The desired latency boundary begins when a question is dispatched and ends when the scored answer is ready. The stored MAB output meets that boundary for \systemname{} because it scores the online answer and retains the same request's latency. The driver launches concurrent requests (concurrency 12 in the runbook), so the value is per-request wall time under load rather than a serial benchmark.

The stored LoCoMo and LongMem outputs do not meet the desired boundary. They preserve latency for an online answer, then score a different offline Stage-3 answer without recording Stage-3 duration. An internal ten-call Sonnet probe reports a 4.22-second mean (median 3.96, p90 7.81), but its raw rows are absent and it does not reconstruct per-question Stage-3 time. We therefore omit cross-system end-to-end latency comparisons for LoCoMo and LongMem rather than pair accuracy and timing from different answer paths.

\subsection{Modeled monetary cost}
Dollar figures are model-based operational budgets, not measured billing totals. The package does not persist complete provider usage fields or include billing exports, and the scored LoCoMo/LongMem Stage-3 path cannot be reconstructed from the old per-query estimates. We therefore omit exact per-query cost comparisons and retain only the coarse per-run budget in Appendix B.

\subsection{Benchmark metrics and judge scope}
The memory-agent task suite uses its canonical macro over accurate retrieval (AR), test-time learning (TTL), long-range understanding (LRU), and selective forgetting (SF). LoCoMo factoid categories use the vendored language-model judge, while the adversarial abstention category uses the benchmark's lexical scorer. LongMemEval uses its task-type-aware binary judge.

No additional human-agreement study was performed. Language-model-judge scores therefore measure agreement with the specified benchmark judge, not human-validated correctness. The LoCoMo prompt originated with one compared system, creating a potential circularity; using the same prompt for all outputs standardizes scoring but does not eliminate that concern. Appendix B states these limits explicitly.

\subsection{Repetition and statistical scope}
The two MAB reference repetitions score 61.72\% and 61.26\%, with arithmetic mean 61.49\% and sample standard deviation 0.32 percentage points. Reporting the runs explicitly avoids making two observations resemble a confidence interval. The other reference-system cells are single-run unless stated otherwise. A local reproduction with retained outputs and a separately reported teammate run appear as consistency checks; neither is pooled with the reference because their provenance, environments, and output snapshots differ.

We report descriptive percentages and percentage-point differences. We do not report p-values, confidence intervals, or claims of statistical significance. Terms such as ``higher,'' ``lower,'' and ``equal at displayed precision'' describe the observed outputs only. Appendix B summarizes this statistical boundary.

\subsection{Router evidence}
The frozen classifier was selected on 1,065 synthetic question-shape examples, reaching 96.06\% overall and 95.80\% macro accuracy. A read-only real-development sanity check on 568 consensus-labeled questions reached 90.14\% micro and 89.37\% macro accuracy. Appendix C reports the per-mode evidence and the runtime overrides.

These data show that the classifier carries question-shape signal. Measuring operational routing efficiency would additionally require a held-out procedure-level confusion matrix and misroute penalties. Stored outputs also lack enough payload provenance to resolve every strategy amendment. The supported claim is therefore implementation of automatic and explicit per-query dispatch, not router optimality or an advantage over fixed policies.

\subsection{Intervention and observational analyses}
Section 6 distinguishes three evidence levels. Configuration-level benchmark comparisons vary complete memory systems. Route-conditioned task and error breakdowns reuse completed outputs and are observational. The reported abstention-on/off comparison is an author-provided configuration difference; the available package contains the with-abstention scored output but not the paired base output, so even this mechanism comparison is not independently auditable as a strict one-variable ablation.

\subsection{Threat from benchmark contamination}
The public benchmarks may occur in the pretraining data of the synthesis and judge models. Sharing model versions across systems reduces one source of variation but does not guarantee that contamination affects all retrieved contexts equally. Absolute scores, and potentially relative differences, should therefore be interpreted with this caveat.

\section{Results}
This section reports completed configuration-level accuracy measurements. The evaluation package includes baseline scores but not the corresponding raw outputs. Timing and cost are discussed separately because the scored LoCoMo and LongMem answers were produced by an offline synthesis path whose latency and token usage were not persisted. We therefore do not present an aligned three-axis comparison.

\subsection{Three-benchmark accuracy summary}

\begingroup
\scriptsize
\setlength{\tabcolsep}{3pt}
\renewcommand{\arraystretch}{1.08}
\begin{longtable}{L{0.38\linewidth} L{0.16\linewidth} L{0.16\linewidth} L{0.16\linewidth}}
\caption{Reported aggregate scores and derived cross-benchmark summaries. Raw baseline rows are unavailable.}\label{tab:aggregate-accuracy}\\
\toprule
Benchmark or summary & Mem0 v2 OSS & \comparatorname{} & \systemname{} \\
\midrule
\endfirsthead
\toprule
Benchmark or summary & Mem0 v2 OSS & \comparatorname{} & \systemname{} \\
\midrule
\endhead
Long-conversational memory, mixed full set (n = 1,986) & 13.08\% & 63.10\% & 81.22\%\textsuperscript{a} \\
Memory-agent task suite (n = 3,671) & 34.03\% & 52.22\% & 61.49\%; runs 61.72\% and 61.26\% \\
Longitudinal personal memory (n = 500) & 24.00\% & 57.00\% & 86.00\% \\
\midrule
Minimum of three benchmark aggregates (``floor'') & 13.08\% & 52.22\% & 61.49\% \\
Unweighted geometric mean & 22.02\% & 57.27\% & 75.45\% \\
\bottomrule
\end{longtable}
\endgroup

\noindent
\begin{minipage}{\linewidth}
\small
\textsuperscript{a} Corrected from 81.17\% by repairing a one-row duplicate-question category collision in the completed scored output; Section 5.4 documents the correction. Baseline values retain the externally reported scorer outputs because raw baseline rows are unavailable. The two MAB observations are shown explicitly; their arithmetic mean rounds to 61.49\%. The minimum and geometric mean equal-weight the three reported benchmark percentages. Because the underlying metrics differ, these are descriptive summaries rather than pooled accuracy, formal robustness statistics, or routing-effect estimates.
\end{minipage}

\begin{center}
\centering
\includegraphics[width=\textwidth]{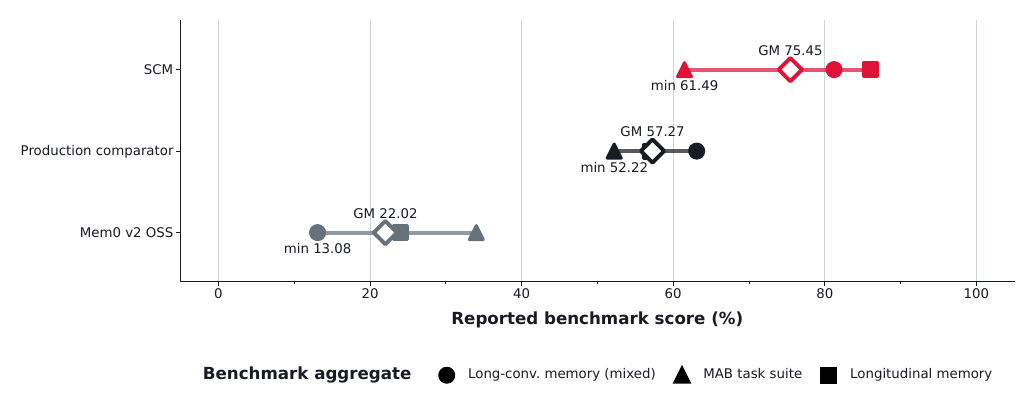}
\captionsetup{type=figure,hypcap=false}
\caption{Cross-workload profiles derived from Table~\ref{tab:aggregate-accuracy}. Benchmark aggregates are shown with distinct point shapes, the leftmost point is each configuration's minimum, and the diamond is its unweighted geometric mean. Lines connect each configuration's observed minimum and maximum; they do not imply continuity or an ordering among benchmarks. The underlying metrics differ, and the summaries are descriptive.}
\label{fig:cross-workload-profile}
\end{center}

The pattern is descriptive. \systemname{} records the largest displayed aggregate on all three benchmarks and the largest observed minimum, but the memory-agent suite's sub-dataset results remain uneven. Because baseline rows are unavailable and most cells are single-run, the numerical differences are not presented as significance-tested wins or as evidence that routing caused the cross-workload profile.

\subsection{Reference and reproduction runs}
Table~\ref{tab:run-provenance} keeps the reported \systemname{} runs separate because they differ in environment and provenance. The author reference is used in Table~\ref{tab:aggregate-accuracy}. The local reproduction is preserved in the repository; the teammate values are externally reported without row-level output in the package.

\begingroup
\scriptsize
\begin{longtable}{L{0.13\linewidth} L{0.13\linewidth} L{0.17\linewidth} L{0.13\linewidth} L{0.15\linewidth} L{0.10\linewidth}}
\caption{Reported \systemname{} runs and their artifact provenance.}\label{tab:run-provenance}\\
\toprule
Run source & Date / build & MAB v3 & LongMem & LoCoMo & Notes \\
\midrule
\endfirsthead
\toprule
Run source & Date / build & MAB v3 & LongMem & LoCoMo & Notes \\
\midrule
\endhead
Author reference & 2026-06-15, deployed v4 fast build & 61.49\% overall; runs 61.72\% and 61.26\% & 86.0\% overall, n = 500 & 84.87\% factoid; 68.61\% abstention; 81.22\% mixed & Reference row used in Table~\ref{tab:aggregate-accuracy} \\
Local reproduction & 2026-06-16 local rep & 62.19\% overall; AR 78.15\%, TTL 50.32\%, LRU 57.94\%, SF 62.38\% & 85.4\% overall & 85.78\% factoid; 68.83\% abstention; 81.97\% mixed & Raw scored output included \\
Teammate report & Externally reported run & 61.43\% overall; AR 78.92\%, TTL 50.33\%, LRU 53.72\%, SF 62.75\% & 86.4\% overall & 84.86\% factoid; 68.46\% abstention; 81.17\% mixed & Original scorer cut; no row-level output \\
\bottomrule
\end{longtable}
\endgroup

The author reference and local reproduction with retained outputs fall within roughly one point of one another on each headline accuracy metric. This is consistency evidence, not a formal reproducibility or variance estimate: only the MAB author reference contains two repetitions, and the environments are not exchangeable. The externally reported values are numerically similar but provide context only because their underlying rows are unavailable.

\subsection{Memory-agent task-suite breakdown}
Table~\ref{tab:mab-breakdown} reports the author-reference competency breakdown. Accurate retrieval (AR), test-time learning (TTL), long-range understanding (LRU), and selective forgetting (SF) were defined in Section 4.5. The sub-dataset aliases are: single- and multi-hop document question answering (SH-Doc-QA and MH-Doc-QA), event question answering (EventQA), LongMemEval-S* (LME-S*), multi-class classification (MCC), movie recommendation (Movie-Rec), InfiniteBench summarization (InfBench-Sum), detective question answering (Detective-QA), and single- and multi-hop fact consolidation (FC-SH and FC-MH). The MCC group averages BANKING77, CLINC150, natural-language understanding (NLU), and the TREC coarse- and fine-label tasks.

\begingroup
\scriptsize
\begin{longtable}{L{0.14\linewidth} L{0.12\linewidth} L{0.56\linewidth}}
\caption{Author-reference MAB competency means across two runs and sub-dataset accuracy from the retained fast-build repetition.}\label{tab:mab-breakdown}\\
\toprule
Competency & Mean & Per-dataset detail \\
\midrule
\endfirsthead
\toprule
Competency & Mean & Per-dataset detail \\
\midrule
\endhead
AR & 78.62\% & SH-Doc-QA 96.0\%; MH-Doc-QA 85.0\%; EventQA 86.3\%; LME-S* 45.7\% \\
TTL & 50.05\% & MCC group 86.40\%; Movie-Rec 13.9\% \\
LRU & 54.67\% & InfBench-Sum 27.5\%; Detective-QA 80.3\% \\
SF & 62.62\% & FC-SH 82.7\%; FC-MH 42.8\% \\
Overall & 61.49\% & Macro over AR, TTL, LRU, and SF \\
\bottomrule
\end{longtable}
\endgroup

The aggregate masks substantial heterogeneity. SH-Doc-QA, MH-Doc-QA, EventQA, and FC-SH are the strongest displayed cells; Movie-Rec, InfBench-Sum, FC-MH, and LME-S* are substantially lower. Route-conditioned failure counts in Section 6 use these cells to formulate engineering hypotheses, not to assign causal credit to a procedure.

\subsection{Long-conversational-memory scoring cuts and one-row correction}
The benchmark contains four factoid categories and one adversarial abstention category. Factoids use the vendored language-model judge; abstention uses the benchmark paper's lexical scorer. We report their scores separately and also give a mixed full-benchmark average for continuity with the supplied evaluation.

The original scoring join keyed duplicated questions by question text. One question---``What did Gina receive from a dance contest?''---appears once as a category-4 factoid with gold answer ``a trophy'' and once as a category-5 adversarial item with gold answer ``undefined.'' The join assigned both scored rows the adversarial metadata, yielding category counts 1,539/447 rather than the fixture's 1,540/446 split. In both repository-backed \systemname{} outputs, the prediction for the displaced factoid is ``A trophy with a glass globe on top,'' an exact-match-safe correct answer. Reclassifying that existing row, without issuing a new model call, gives:

\begin{longtable}{L{0.36\linewidth} L{0.18\linewidth} L{0.18\linewidth} L{0.18\linewidth}}
\caption{Corrected LoCoMo scoring cuts for repository-backed \systemname{} outputs.}\label{tab:locomo-corrected}\\
\toprule
Run & Factoid categories (n = 1,540) & Abstention (n = 446) & Mixed full set (n = 1,986) \\
\midrule
\endfirsthead
\toprule
Run & Factoid categories (n = 1,540) & Abstention (n = 446) & Mixed full set (n = 1,986) \\
\midrule
\endhead
Author reference & 84.87\% & 68.61\% & 81.22\% \\
Local reproduction & 85.78\% & 68.83\% & 81.97\% \\
\bottomrule
\end{longtable}

The mixed score combines two different metrics and should not be interpreted as a homogeneous accuracy measure. We use it only as an end-to-end summary and give the component cuts alongside it. When a comparison refers specifically to \systemname{} factoid recall, the relevant result is 84.87\%; 81.22\% is reserved for the explicitly labeled mixed full set. We do not compare 84.87\% directly with a supplied baseline unless that baseline is confirmed to use the same factoid cut. Because raw production-comparator and Mem0 rows are absent, their supplied values have not been reprocessed through this one-row correction.

\subsection{Reported abstention configuration difference}
The author-provided comparison reports 2.46\% on the original 447-row adversarial cut with entity-grounded abstention disabled and 68.46\% with it enabled, a 66.00-point difference under that original scorer output. The enabled reference output is present and becomes 68.61\% after the category correction above. The paired disabled output is absent, so we cannot independently verify that the entity-grounding toggle was the only difference or recompute its corrected denominator. We therefore treat the 66-point change as a reported configuration difference, not an audited causal ablation.

In the enabled factual path, the synthesis prompt instructs the model to identify the question's main entity and answer ``No information available'' when that entity is unsupported by retrieved memories. This is model-directed prompt behavior, not a deterministic post-retrieval filter. The missing disabled output prevents a mechanism-level effect estimate.

\subsection{Longitudinal personal-memory task types}

\begin{longtable}{L{0.30\linewidth} L{0.08\linewidth} L{0.17\linewidth} L{0.17\linewidth} L{0.17\linewidth}}
\caption{Reported LongMemEval accuracy by task type.}\label{tab:longmem-task-types}\\
\toprule
Task type & n & Mem0 v2 OSS & Comparator & \systemname{} \\
\midrule
\endfirsthead
\toprule
Task type & n & Mem0 v2 OSS & Comparator & \systemname{} \\
\midrule
\endhead
Single-session-user & 70 & 34.29\% & 77.14\% & 95.71\% \\
Single-session-assistant & 56 & 5.36\% & 94.64\% & 94.64\% \\
Single-session-preference & 30 & 66.67\% & 53.33\% & 83.33\% \\
Multi-session & 133 & 31.58\% & 53.38\% & 78.95\% \\
Temporal-reasoning & 133 & 15.04\% & 31.58\% & 83.46\% \\
Knowledge-revision & 78 & 14.10\% & 62.82\% & 88.46\% \\
\bottomrule
\end{longtable}

The observed \systemname{}--comparator differences are +18.57 points on single-session-user, 0 on single-session-assistant, +30.00 on single-session-preference, +25.57 on multi-session, +51.88 on temporal-reasoning, and +25.64 on knowledge-revision. The completed local trace records mode labels, not the final executed procedure. Temporal questions are distributed mainly across time-anchored, long-form, and single-fact labels; knowledge-update questions are mostly labeled single-fact, with smaller long-form, time-anchored, and latest-version groups. No removal ablations were run. The task scores therefore cannot be assigned to a unique procedure.

\begin{center}
\centering
\includegraphics[width=\textwidth]{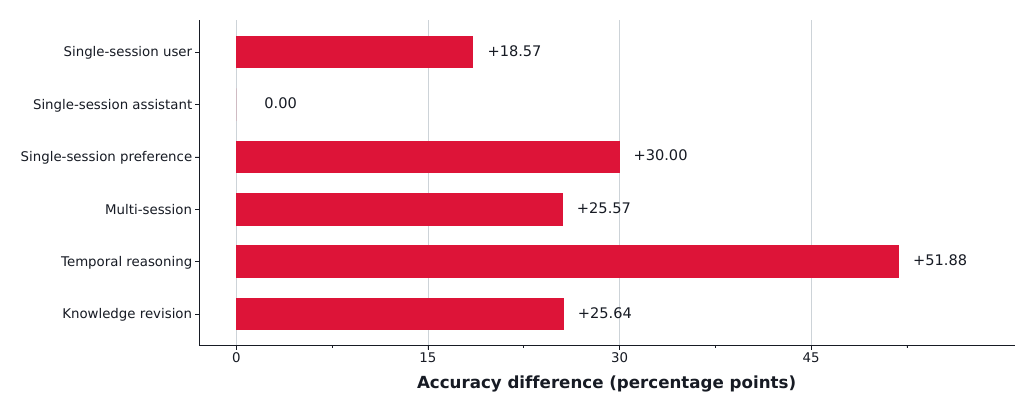}
\captionsetup{type=figure,hypcap=false}
\caption{Observed longitudinal task-type score differences relative to the supplied comparator values. These descriptive differences should not be read as executed-procedure or causal effects.}
\label{fig:longmem-task-deltas}
\end{center}

\subsection{Operational timing and cost scope}
MAB uses the answer returned by the online evaluation route, so its stored request latency and scored answer refer to the same path. The reference reports 9.00 seconds per request under the benchmark launch configuration. The full-run driver uses concurrent requests (the runbook uses concurrency 12), so this is per-request wall time under load rather than a serial single-thread benchmark.

LoCoMo and LongMem use a separate offline Stage-3 synthesizer for the scored answer. Their retrieval records retain latency for the earlier online answer and do not record Stage-3 call duration. Pairing those latencies with Stage-3 accuracy would mix two answer paths, so we omit end-to-end latency claims for these benchmarks. The same missing Stage-3 token records prevent a per-query cost comparison aligned to the scored answers. Appendix B reports only the coarse runbook budget.

These evidence boundaries support a limited conclusion: the completed runs measure one implemented routed configuration and expose task- and mode-conditioned hypotheses. Comparisons against a fixed procedure, aligned end-to-end latency across systems, and component-level attribution require additional controls.

\section{Diagnostic Analysis and Limitations}

\subsection{Evidence levels}

The retained package supports three levels of interpretation. First, benchmark tables compare complete configurations; many components vary together and raw baseline rows are unavailable. Second, \systemname{} traces retain semantic mode labels, retrieved chunks, and scored outcomes, enabling reproducible mode-conditioned diagnostics. They do not always retain the final procedure after runtime gates and fallbacks. Third, an author-provided abstention comparison changes a complete configuration, but the disabled per-question output is absent. Consequently, the evidence supports description of the routed configuration, not a comparison of automatic routing against a fixed policy.

\subsection{Patterns in completed traces}

Table~\ref{tab:mode-conditioned-patterns} identifies high-mass error strata in the repository-backed local reproduction. Counts are useful for selecting cases to inspect, but they are not causal effects and should be interpreted alongside each route's prevalence.

\begin{table}[htbp]
\small
\centering
\begin{tabular}{L{0.31\linewidth} L{0.28\linewidth} L{0.31\linewidth}}
\toprule
Observed stratum & Largest recorded mode group & Diagnostic reading \\
\midrule
LongMem temporal failures & Long-form (14/25), then time-anchored (9/25) & Errors cluster in two semantic routes \\
LongMem knowledge-update failures & Single-fact (5/10) & Most failures are not labeled latest-version \\
MAB FC-MH failures & Long-form (214/229) & Relation-chain errors concentrate in this label \\
MAB Movie-Rec failures & Single-fact (179/179) & Recommendation questions are poorly served by the recorded route \\
MAB EventQA failures & Long-form (188/191) & Event errors concentrate in this label \\
LoCoMo abstention failures & Single-fact (110/139) & Residual refusal errors concentrate in lookup \\
\bottomrule
\end{tabular}
\caption{Mode-conditioned failure strata from completed outputs. The final executed procedure is not always known.}
\label{tab:mode-conditioned-patterns}
\end{table}

LongMemEval provides the clearest retrieval-versus-synthesis diagnostic. All 73 judged failures in the local 85.4\% reproduction have \texttt{hit\_at\_k=true}. The largest residual cells are multi-session (26/133 failures) and temporal reasoning (25/133). Representative rows contain evidence-use errors: incomplete counting despite several relevant mentions, generic advice instead of a stored preference, incorrect ordering of retrieved events, and abstention when current and prior facts are present. The hit flag is coarse, but the pattern is consistent with errors after initial retrieval.

MAB failures are similarly concentrated. FC-MH records 229/400 failures, Movie-Rec 179/200 recall misses, EventQA 191/1,500 failures, and LME-S* 164/300 failures. These strata motivate relation-chain, recommendation, and event-synthesis investigation; they do not identify a final executed procedure. On the corrected LoCoMo adversarial cut, the local reproduction has 139/446 residual failures. Many are near-miss refusals or semantically reasonable corrections that fail the lexical target, reinforcing the decision to report factoid and abstention metrics separately.

The author-provided abstention comparison reports 2.46\% with entity-grounded abstention disabled and 68.46\% with it enabled on the original 447-row adversarial cut. The enabled output is present and becomes 68.61\% after the category correction; the disabled output is not present. Source inspection finds a prompt instruction to emit ``No information available'' when the main entity is unsupported, not a deterministic entity filter. The prompt and reported configuration difference are therefore documented separately rather than combined into a mechanism-level effect estimate.

\subsection{Threats to validity}

\begin{itemize}
\item \textbf{Baseline provenance.} Raw Mem0 and production-comparator predictions are unavailable, so prompt parity, row-level corrections, and paired uncertainty cannot be audited.
\item \textbf{Answer-path alignment.} MAB scores the online response whose latency is stored. LoCoMo and LongMemEval score offline Stage-3 synthesis while retaining timing for an earlier answer; exact Stage-3 usage is absent.
\item \textbf{Repetition and judges.} The MAB reference has two repetitions and most other cells are single-run. Language-model judges may be prompt-, verbosity-, and contamination-sensitive, and no additional human-agreement study was performed.
\item \textbf{System non-invariants.} Mem0 uses ChromaDB, while \systemname{} and the production comparator use pgvector. Generator, retrieval, storage, and prompt differences remain part of the complete-configuration comparison.
\item \textbf{Routing provenance.} Stored rows retain semantic labels but not every post-gate tier, graph strategy, fallback, or prompt family. The active graph field and amendment note also disagree, and the real-development router set contains almost no latest-version examples.
\item \textbf{Scoring correction.} A question-text join displaced one LoCoMo factoid into the adversarial cut. The existing \systemname{} rows are corrected by fixture identity, but raw baselines cannot be reprocessed in the same way.
\end{itemize}

\subsection{Scope and next steps}

The available evidence supports an implementation claim and a descriptive characterization of one routed configuration. Claims about routing efficiency, component causality, statistical significance, or human-validated judge accuracy require new controls. The highest-value next steps are to persist complete routing provenance, align scored answers with timing and token usage, compare routed and fixed procedures on identical corpus snapshots, and audit a stratified judge sample. Those controls remain outside the claims of this version.

\section{Conclusion}

Supra Cognitive Modes implements a per-query control interface over a shared agent-memory substrate. Explicit or automatically selected semantic modes map to retrieval and synthesis payloads, while runtime gates can dispatch among direct, graph-capable, and long-form procedures without rebuilding the memory store. Mandatory ingest creates a queryable lexical, embedding, and relation substrate; asynchronous supplements add nullable signals for selected procedures.

Across three reported benchmark comparisons, the deployed configuration records the largest aggregate accuracy and the local reproduction with retained outputs yields similar headline scores. The rows also expose concrete weaknesses in recommendation, long-range summarization, multi-hop conflict resolution, temporal evidence use, and lexical abstention. Because the comparisons are configuration-level, they provide neither a routing-efficiency estimate nor component-level attribution.

The next evaluation step is therefore not another broad benchmark table, but aligned and attributable measurement: persist the final procedure and fallback path, score the same answer whose latency and usage are retained, compare routed and fixed policies on identical snapshots, and validate a stratified sample of judge decisions. Until then, the central result is the implemented mode--payload--procedure interface and its evidence-bounded characterization over one shared substrate.

\bibliography{refs}
\appendix
\section{Methodology details}
This appendix records the executable configuration that is materialized in the reproduction artifact. The goal is to make the reported measurements traceable to files, command entrypoints, and pinned runtime parameters rather than to prose descriptions alone.

\subsection{Artifact sources of truth}

\begin{longtable}{L{0.27\linewidth} L{0.30\linewidth} L{0.35\linewidth}}
\caption{Artifact sources of truth.}\label{tab:artifact-sources-of-truth}\\
\toprule
Method component & Artifact path & What the artifact fixes \\
\midrule
\endfirsthead
\toprule
Method component & Artifact path & What the artifact fixes \\
\midrule
\endhead
Question-shape classifier & \path{supraos-memory-benchmark/scripts/classifier-rebuild/frozen-classifier.json} & Classifier prompt, model, temperature, mode set, synthetic-selection scores, and dev-sanity scores. \\
Mode-to-runtime payload map & \path{supraos-memory-benchmark/scripts/classifier-rebuild/phaseB-mode-config.json} & Retrieval config selected per mode, final \texttt{top\_k}, synth mode, force-tier flags, and amendment notes. \\
Retrieval configs & \path{evals/memory-agent-bench/configs/*.json} & Retriever mix, BM25 parameters, RRF constant, post-processing stack, and disabled/non-functional comment keys. \\
Full-run retrieval driver & \texttt{scripts/phaseB-cm-}\newline\texttt{fullrun-retrieval.py} & Per-question owner-address mapping, reference-date construction, classifier invocation, runtime fallbacks, checkpointing, and output provenance fields. \\
Offline synthesis driver & \path{scripts/run_memory_pipeline.py}; \path{scripts/memory_system.py} & Local Sonnet synthesis path used for LoCoMo and LongMem scoring, role cache, intent classifier, prompt family, and max-token policy. \\
Benchmark scorers & \path{scripts/score_mab.sh}; \path{scripts/score_locomo.sh}; \path{scripts/score_longmem.sh} & End-to-end scoring entrypoints and the ordered conversion steps required before judging. \\
Upstream judge prompts & \path{scripts/scoring/upstream_prompts/*.txt} & Byte-vendored prompts for MemoryAgentBench LRU and the Mem0 LoCoMo judge. \\
Question fixtures & \path{data/gbrain-rerun-2026-05-16/*.ndjson} & The 6,157 benchmark questions used by the retrieval driver. \\
\bottomrule
\end{longtable}

\subsection{Pinned model and call parameters}

\begin{longtable}{L{0.30\linewidth} L{0.62\linewidth}}
\caption{Pinned model and call parameters.}\label{tab:pinned-model-call-parameters}\\
\toprule
Component & Pinned value in the artifact \\
\midrule
\endfirsthead
\toprule
Component & Pinned value in the artifact \\
\midrule
\endhead
Frozen routing classifier & \texttt{claude-haiku-4-5-20251001}, temperature 0, max tokens 16. \\
Offline role and intent classification & \texttt{claude-haiku-4-5-20251001}, temperature 0, max tokens 30. \\
Offline synthesis & \texttt{claude-sonnet-4-5-20250929}, temperature 0. Max tokens are 4,096 for summaries, 1,024 for recommendation lists, and 600 for other answers. \\
DetQA resynthesis & \texttt{claude-sonnet-4-5-20250929}, temperature 0, max tokens 1,024, canonical MemoryAgentBench detective prompt plus hardening clause. \\
MAB LRU summarization judge & \texttt{gpt-4o-2024-05-13}, temperature 0.1, max tokens 4,096, three prompt calls per InfBench-Sum question. \\
LoCoMo judge & \texttt{gpt-4o-mini-2024-07-18}, temperature 0, max tokens 300, JSON response format, categories 1--4 only. \\
LongMemEval judge & \texttt{gpt-4o-2024-08-06}, temperature 0, max tokens 10, task-type-specific prompt. \\
OpenAI helper retry policy & Six attempts, exponential backoff, 120-second timeout. \\
Anthropic helper retry policy & Six attempts, exponential backoff, 180-second timeout. \\
\bottomrule
\end{longtable}

\subsection{Retrieval and scoring parameters}

\begin{longtable}{L{0.30\linewidth} L{0.62\linewidth}}
\caption{Retrieval and scoring parameters.}\label{tab:retrieval-scoring-parameters}\\
\toprule
Parameter & Value fixed by code or config \\
\midrule
\endfirsthead
\toprule
Parameter & Value fixed by code or config \\
\midrule
\endhead
Embedding provider & \texttt{openai-large}, corresponding to OpenAI \texttt{text-embedding-3-large} at 1,024 dimensions in the runbook and ingest documentation. \\
Final retrieval depth & \texttt{top\_k = 100} in the \texttt{/api/memory/eval-run} body for the benchmark runs. Retrieval-config candidate fetch depths may be larger before fusion and post-processing. \\
BM25 retriever & Okapi BM25 with \texttt{k1 = 1.5}, \texttt{b = 0.75}. Canonical config uses \texttt{top\_k = 100}; h17/h40 configs use wider candidate pools. \\
Dense retrievers & Chunk, sentence, and paragraph embedding retrievers in \texttt{db\_ann} mode. \\
Fusion & Reciprocal-rank fusion with \texttt{k\_constant = 30}. \\
Recency booster & Present in h17/h40 configs with boost factor 0, because all chunks in the frozen paper-grade run are freshly ingested and a wall-clock recency boost would be non-deterministic. \\
Narrative rerank & Enabled in h17/h40 configs with \texttt{boost\_factor = 0.5}, \texttt{location\_weight = 0.5}, and \texttt{min\_freq = 2}; the driver rewrites the \texttt{ingest\_run\_id} at runtime. \\
Driver concurrency & The runbook commands use concurrency 12 for full benchmark runs. The retrieval driver also defines a hard cap of 24 concurrent \texttt{eval-run} POSTs. \\
MAB canonical scoring & Overall is the mean of AR, TTL, LRU, and SF. TTL is the mean of in-context classification and recommendation recall; LRU is the mean of InfBench-Sum F1 and Detective-QA substring exact match after DetQA resynthesis. \\
LoCoMo scoring & Headline uses the Mem0 lenient LLM judge for categories 1--4; category 5 is reported separately with the benchmark paper's abstention lexical rule. \\
LongMemEval scoring & Official task-type-aware binary judge, with abstention prompts selected for abstention question IDs. \\
\bottomrule
\end{longtable}

\subsection{Reproducibility guardrails}

Two guardrails are methodologically important enough to call out. First, \path{scripts/bolt_from_stage3.py} joins Stage 3 synthesis output back to source rows by question ID rather than by parallel-completion order. The runbook records that an enumerate-based join measured LongMem at 21.8\%, while the corrected question-ID join measured 88.2\% on the same rep. Second, the MAB scorer applies the Detective-QA resynthesis before computing the canonical LRU macro; without this step, Detective-QA rows are judged against summary-style predictions rather than the benchmark's detective prompt.

The current artifact snapshot is not fully portable: \path{scripts/scoring/score_locomo.py} and \path{scripts/scoring/score_longmem.py} use hard-coded absolute roots instead of deriving \texttt{ROOT} from their own file paths. This does not change the recorded results, but those scripts require path normalization before they can be reused in another environment.

\section{Reproducibility and Evidence Scope}

\subsection{Judge-based metrics}

LoCoMo factoids, LongMemEval task types, and the MemoryAgentBench long-range metric use benchmark-provided or comparator-derived language-model judge prompts. Exact prompts, model identifiers, temperatures, and output limits are recorded in Appendix A. No additional human-agreement annotation was performed. Reported values therefore measure agreement with the specified judge pipelines rather than independently validated correctness. This distinction matters for semantically plausible answers, verbosity-sensitive cases, and LoCoMo factoids, where the prompt originated with one compared system.

\subsection{Comparator scope}

The Mem0 reference is described as using \texttt{mem0ai==2.0.3}, \texttt{top\_k=100}, language-model reranking with \texttt{gpt-4o-mini}, and Claude Haiku 4.5 extraction. The production comparator is treated as an evaluated configuration rather than a fully inspectable research artifact. The package contains aggregate baseline values and the reproduction command surface, but not raw baseline predictions, retrieval traces, complete logs, or billing records. Baseline comparisons are therefore reported values rather than independently recomputed measurements.

\subsection{Cost and statistical reporting}

The runbook contains planning estimates of approximately USD~172 and two to three hours for one full \systemname{} reproduction, including provider calls and judges. These are budget estimates, not observed bills: provider token usage and reconciled billing exports are not persisted consistently, and the LoCoMo and LongMemEval Stage-3 synthesis path cannot be reconstructed from earlier per-query estimates. The paper consequently reports no exact cross-system per-query cost.

The MAB reference repetitions are 61.72\% and 61.26\%, with arithmetic mean 61.49\% and sample standard deviation 0.32 percentage points. The tables show both observations instead of presenting a mean with a plus-or-minus standard deviation. Other reference cells are single-run unless stated otherwise. Raw baseline rows and sufficient independent repetitions are unavailable, so the manuscript reports no p-values, confidence intervals, or formal effect sizes. Percentage-point differences are descriptive arithmetic on the retained scores.

\subsection{Artifact scope}

The current artifact snapshot contains fixtures, scripts, configurations, prompts, and \systemname{} outputs, but does not include row-level baseline outputs, the abstention-disabled output, complete token ledgers, probe rows, human annotations, or significance-test outputs. Credentials, private-infrastructure references, nonportable paths, and third-party licensing require review before distribution. No public artifact accompanies this version; a later release would require sanitization, licensing review, portable path configuration, and a file manifest linking every reported result to its source artifact.

\section{Classifier routing accuracy}
This appendix documents the routing evidence present in the repository. It distinguishes three layers that can otherwise be conflated: the frozen four-mode classifier, the mode-to-runtime payload mapping, and post-classifier runtime overrides that are visible in the full-run driver.

Two classifier copies are present. The nested \path{supraos-memory-benchmark} package uses the semantic names reported in the benchmark outputs; the top-level legacy copy uses \texttt{DIRECT\_RECALL}, \texttt{LONG\_RANGE\_READER}, \texttt{TEMPORAL\_RECALL}, and \texttt{SKEPTIC}. This appendix cites the nested package explicitly. The copies are aliases from different packaging revisions and should not be treated as byte-identical provenance.

\subsection{Frozen classifier artifact}

\begin{longtable}{L{0.33\linewidth} L{0.57\linewidth}}
\caption{Frozen classifier artifact fields.}\label{tab:frozen-classifier-artifact}\\
\toprule
Field & Value \\
\midrule
\endfirsthead
\toprule
Field & Value \\
\midrule
\endhead
Artifact & \path{supraos-memory-benchmark/scripts/classifier-rebuild/frozen-classifier.json}. \\
Frozen at & 2026-05-17T17:46:24+0800. \\
Model & \texttt{claude-haiku-4-5-20251001}. \\
Temperature & 0. \\
Winner candidate & \texttt{c5-balanced}. \\
Frozen modes & \texttt{SINGLE\_FACT\_LOOKUP}, \texttt{LONG\_FORM\_SYNTHESIS}, \texttt{TIME\_ANCHORED\_LOOKUP}, and \texttt{LATEST\_VERSION\_RESOLUTION}. \\
Selection set & 1,065 synthetic shape-based questions, described in the artifact as off-test and independently verified. \\
Selection criterion & Highest macro accuracy across five candidate prompts. \\
Synthetic overall / macro & 96.06\% overall, 95.80\% macro. \\
Synthetic per-mode accuracy & Single-fact lookup 91.73\%, long-form synthesis 99.72\%, time-anchored lookup 92.24\%, latest-version resolution 99.54\%. \\
Real-dev sanity check & 616-question read-only slice; 568 consensus-labeled questions after two strong labelers. \\
Real-dev agreement & Inter-labeler agreement 93.88\%; classifier micro 90.14\%, macro 89.37\% on the consensus subset. \\
Real-dev per-mode accuracy & Single-fact lookup 88.13\% (n = 278), long-form synthesis 95.10\% (n = 204), time-anchored lookup 84.88\% (n = 86). \\
Known limitation & The dev slice contains almost no verbalized-conflict questions, so \texttt{LATEST\_VERSION\_RESOLUTION} is validated synthetically but not measured on the real-dev slice. \\
\bottomrule
\end{longtable}

The classifier always returns one of the four frozen modes after parsing; unparseable outputs default to \texttt{SINGLE\_FACT\_LOOKUP}. The full prompt is stored verbatim in the frozen classifier artifact rather than reprinted here.

\subsection{Mode-to-runtime mapping}

{\footnotesize
\begin{longtable}{L{0.30\linewidth} L{0.22\linewidth} L{0.38\linewidth}}
\caption{Mode-to-runtime mapping.}\label{tab:mode-runtime-mapping}\\
\toprule
Runtime mode & Retrieval config & Eval-run flags fixed by the artifact \\
\midrule
\endfirsthead
\toprule
Runtime mode & Retrieval config & Eval-run flags fixed by the artifact \\
\midrule
\endhead
\texttt{SINGLE\_FACT\_LOOKUP} & \texttt{h40-temporal-synth} & \texttt{force\_tier=simple}, \texttt{top\_k=100}, \texttt{synthesizer=sonnet}; payload also sets \texttt{synth\_mode} to this mode. \\
\texttt{TIME\_ANCHORED\_LOOKUP} & \texttt{h40-temporal-synth} & \texttt{force\_tier=simple}, \texttt{top\_k=100}, \texttt{synthesizer=sonnet}; payload also sets \texttt{synth\_mode} to this mode. \\
\texttt{LATEST\_VERSION\_RESOLUTION} & \texttt{h40-temporal-synth} & \texttt{force\_tier=simple}, \texttt{top\_k=100}, \texttt{synthesizer=sonnet}; payload also sets \texttt{synth\_mode} to this mode. \\
\texttt{LONG\_FORM\_SYNTHESIS} & \texttt{h17-narrative} & \texttt{multi\_hop\_strategy}: \texttt{graph\_routed} in the JSON field, with an amendment note to \texttt{graph\_then\_iterative}; \texttt{graph\_planner\_model}: \texttt{haiku}; \texttt{top\_k=100}; \texttt{synthesizer=sonnet}; payload also sets \texttt{synth\_mode} to this mode. \\
\texttt{ICL\_BROAD\_CONTEXT} & \texttt{h40-temporal-synth} & Runtime route added in the mode config for in-context-learning corpora; uses simple tier, \texttt{top\_k=100}, and single-fact lookup payload metadata. \\
\bottomrule
\end{longtable}
}

The inspected synthesizer source accepts a \texttt{synth\_mode} field but does not branch directly on the frozen classifier labels. Prompt selection in the inspected code is driven by intent, question format, and long-form synthesis flags. The \texttt{synth\_mode} values above should therefore be treated as payload metadata unless the final run uses a source version that consumes them directly.

The active mode-config file also carries an amendment note about changing \texttt{LONG\_FORM\_SYNTHESIS} from \texttt{graph\_routed} to \texttt{graph\_then\_iterative}. The active JSON field still reads \texttt{graph\_routed}, and the stored launch command contains no override. Per-question outputs omit the final strategy. We retain this mismatch as a provenance limitation and do not identify either strategy as the one responsible for the reported scores.

\subsection{Runtime overrides and observed routing distribution}

The retrieval driver can alter or supplement the frozen classifier's output using production-observable signals only:

\begin{itemize}
\item A chain-shape detector can force long-form-synthesis routing before the Haiku classifier when the question text matches multi-hop chain patterns.
\item A second chain-shape gate can force \texttt{force\_tier} to \texttt{multi\_hop} in the final \path{/api/memory/eval-run} payload when the query has at least two chain markers.
\item An ICL corpus-shape detector can reroute to \texttt{ICL\_BROAD\_CONTEXT} after inspecting retrieved chunks for repeated \texttt{label:} examples.
\item Empty-retrieval fallbacks can retry an ICL route or use summary-tier retrieval to recover chunks.
\end{itemize}

These rules explain why the full-run output can contain five \texttt{classified\_mode} values even though the frozen Haiku classifier has four modes. In the completed MAB rep stored at \texttt{data/mab-rep1-2026-06-09}, the observed distribution was:

\begin{longtable}{L{0.35\linewidth} L{0.15\linewidth} L{0.20\linewidth}}
\caption{Observed MAB classified-mode distribution.}\label{tab:mab-classified-mode-distribution}\\
\toprule
Observed \texttt{classified\_mode} & Count & Share of 3,671 questions \\
\midrule
\endfirsthead
\toprule
Observed \texttt{classified\_mode} & Count & Share of 3,671 questions \\
\midrule
\endhead
\texttt{LONG\_FORM\_SYNTHESIS} & 2,192 & 59.71\% \\
\texttt{SINGLE\_FACT\_LOOKUP} & 867 & 23.62\% \\
\texttt{ICL\_BROAD\_CONTEXT} & 496 & 13.51\% \\
\texttt{TIME\_ANCHORED\_LOOKUP} & 73 & 1.99\% \\
\texttt{LATEST\_VERSION\_RESOLUTION} & 43 & 1.17\% \\
\bottomrule
\end{longtable}

\subsection{Validation boundary}

The repository does not contain a held-out procedure-level gold assignment, a runtime confusion matrix, per-procedure precision and recall, or measured misroute penalties. The synthetic selection set and consensus real-development slice show that the frozen semantic classifier carries signal, but they do not validate the full classifier-plus-gates execution path. This paper therefore makes no operational router-accuracy or routing-efficiency claim.

\section{Dataset Statistics and Trace Audit}
This appendix reports dataset and run-shape statistics derived from bundled fixtures and completed \systemname{} outputs. It also records evidence limitations for the vector-store comparison and internal synthesis-latency probe.

\subsection{Benchmark fixture sizes}

\begin{longtable}{L{0.22\linewidth} L{0.18\linewidth} L{0.18\linewidth} L{0.34\linewidth}}
\caption{Benchmark fixture sizes.}\label{tab:benchmark-fixture-sizes}\\
\toprule
Benchmark & Question count & Source fixture & Additional corpus statistic available in repo \\
\midrule
\endfirsthead
\toprule
Benchmark & Question count & Source fixture & Additional corpus statistic available in repo \\
\midrule
\endhead
Memory-agent task suite & 3,671 & \path{phaseA-mab.ndjson} & Upstream parquet files and source maps are bundled; the completed MAB rep records 100 retrieved chunks per question. \\
Long-conversational-memory & 1,986 & \path{phaseA-locomo.ndjson}; \path{locomo10.json} & 10 conversation samples, 1,986 QA items, 272 parsed sessions, and 5,882 message-like conversation entries. \\
Longitudinal personal-memory & 500 & \path{phaseA-longmem.ndjson}; \path{longmemeval_oracle.json} & 500 question records, 948 haystack sessions, and 10,960 message-like haystack entries. \\
\bottomrule
\end{longtable}

The question counts are the exact row counts consumed by the full-run retrieval driver. LoCoMo and LongMem corpus counts are computed from the bundled JSON fixtures. The memory-agent task suite's upstream parquet files are included, but the package does not include a mechanically generated document and token-count table for that corpus.

\subsection{MAB sub-dataset distribution}

The completed MAB rep at \texttt{data/mab-rep1-2026-06-09} records the following sub-dataset distribution after tagging:

\begin{longtable}{L{0.32\linewidth} L{0.15\linewidth} L{0.35\linewidth}}
\caption{MAB sub-dataset distribution.}\label{tab:mab-subdataset-distribution}\\
\toprule
Sub-dataset & Count & Scoring role \\
\midrule
\endfirsthead
\toprule
Sub-dataset & Count & Scoring role \\
\midrule
\endhead
EventQA & 1,500 & Accurate-retrieval macro component. \\
FC-MH & 400 & Conflict-resolution / compositional-factoid component. \\
FC-SH & 400 & Conflict-resolution / single-hop fact-consolidation component. \\
LME-S* & 300 & Accurate-retrieval macro component. \\
Movie-Rec & 200 & Recommendation component of TTL. \\
BANKING77 & 100 & In-context classification component of TTL. \\
CLINC150 & 100 & In-context classification component of TTL. \\
InfBench-Sum & 100 & LRU summarization component, judged by three-prompt GPT-4o F1. \\
MH-Doc-QA & 100 & Accurate-retrieval macro component. \\
NLU & 100 & In-context classification component of TTL. \\
SH-Doc-QA & 100 & Accurate-retrieval macro component. \\
TREC-Coarse & 100 & In-context classification component of TTL. \\
TREC-Fine & 100 & In-context classification component of TTL. \\
Detective-QA & 71 & LRU detective component after canonical DetQA resynthesis. \\
\bottomrule
\end{longtable}

\subsection{Recorded mode distributions}

Appendix C reports the MAB distribution. The existing local LoCoMo and LongMem retrieval traces yield the following semantic-mode counts without new model calls:

\begin{longtable}{L{0.30\linewidth} L{0.24\linewidth} L{0.24\linewidth}}
\caption{Recorded semantic-mode distributions.}\label{tab:recorded-mode-distributions}\\
\toprule
Recorded mode & LoCoMo (n = 1,986) & LongMem (n = 500) \\
\midrule
\endfirsthead
\toprule
Recorded mode & LoCoMo (n = 1,986) & LongMem (n = 500) \\
\midrule
\endhead
\texttt{SINGLE\_FACT\_LOOKUP} & 1,286 (64.75\%) & 301 (60.20\%) \\
\texttt{TIME\_ANCHORED\_LOOKUP} & 475 (23.92\%) & 121 (24.20\%) \\
\texttt{LONG\_FORM\_SYNTHESIS} & 222 (11.18\%) & 74 (14.80\%) \\
\texttt{LATEST\_VERSION\_RESOLUTION} & 3 (0.15\%) & 4 (0.80\%) \\
\bottomrule
\end{longtable}

These are recorded classifier labels, not final procedure counts. Runtime gates and fallbacks can change execution after the label is written.

\subsection{Existing-output error-analysis frame}

The completed 2026-06-16 local reproduction contains enough per-question data to support the diagnostic analysis in Section 6 without running a new benchmark condition. The joined fields are: question text, gold answer, prediction, benchmark task type or category, scorer pass/fail field, classified mode, latency, and retrieved chunks. For LongMemEval, the retrieval trace also records \texttt{hit\_at\_k}. File names in the table are relative to the corresponding \path{codex-*-rep1-20260616} output directory unless otherwise noted.

\begin{longtable}{L{0.25\linewidth} L{0.27\linewidth} L{0.20\linewidth} L{0.20\linewidth}}
\caption{Files and joins used for existing-output error analysis.}\label{tab:error-analysis-files-and-joins}\\
\toprule
Benchmark & Scored-output file & Retrieval-trace file & Join convention \\
\midrule
\endfirsthead
\toprule
Benchmark & Scored-output file & Retrieval-trace file & Join convention \\
\midrule
\endhead
Memory-agent task suite & \path{scored-mab-lexical.json}; \path{scored-mab-lru.json} & \texttt{mab-retrieval-}\newline\texttt{tagged.ndjson} & \texttt{\_idx} row index. \\
Long-conversational-memory & \path{results/scored-locomo.json} & \path{locomo-retrieval.ndjson} & scorer row order / \texttt{\_idx}. \\
Longitudinal personal-memory & \path{results/scored-longmem.json} & \texttt{longmem-}\newline\texttt{retrieval.ndjson} & \texttt{question\_id}. \\
\bottomrule
\end{longtable}

The following table gives the scored-failure strata used in Section 6. For LoCoMo categories 1--4, failure means the Mem0-style language-model judge label is not correct. For LoCoMo category 5, failure means the benchmark paper's lexical abstention scorer does not return full credit. For LongMemEval, failure means the task-type binary judge returns 0. For MAB rows, failure means the row's headline metric is 0 for binary or recall-style metrics; InfBench-Sum is reported as a continuous summarization-F1 score rather than converted to a binary failure count.

\begin{longtable}{L{0.27\linewidth} L{0.18\linewidth} L{0.15\linewidth} L{0.30\linewidth}}
\caption{Scored-failure strata in completed outputs.}\label{tab:scored-failure-strata}\\
\toprule
Benchmark stratum & Scored failures & Denominator & Largest associated route strata \\
\midrule
\endfirsthead
\toprule
Benchmark stratum & Scored failures & Denominator & Largest associated route strata \\
\midrule
\endhead
LoCoMo category 1 & 41 & 282 & 30 \texttt{SINGLE\_FACT\_LOOKUP}; 10 \texttt{LONG\_FORM\_SYNTHESIS}. \\
LoCoMo category 2 & 58 & 321 & 49 \texttt{TIME\_ANCHORED\_LOOKUP}. \\
LoCoMo category 3 & 26 & 96 & 15 \texttt{SINGLE\_FACT\_LOOKUP}; 11 \texttt{LONG\_FORM\_SYNTHESIS}. \\
LoCoMo category 4 & 94 & 841 & 56 \texttt{SINGLE\_FACT\_LOOKUP}; 21 \texttt{LONG\_FORM\_SYNTHESIS}; 17 \texttt{TIME\_ANCHORED\_LOOKUP}. \\
LoCoMo category 5 & 139 & 446 & 110 \texttt{SINGLE\_FACT\_LOOKUP}; 19 \texttt{TIME\_ANCHORED\_LOOKUP}. \\
LongMemEval multi-session & 26 & 133 & 13 \texttt{SINGLE\_FACT\_LOOKUP}; 9 \texttt{TIME\_ANCHORED\_LOOKUP}; all 26 have \texttt{hit\_at\_k=true}. \\
LongMemEval temporal-reasoning & 25 & 133 & 14 \texttt{LONG\_FORM\_SYNTHESIS}; 9 \texttt{TIME\_ANCHORED\_LOOKUP}; all 25 have \texttt{hit\_at\_k=true}. \\
LongMemEval knowledge-update & 10 & 78 & 5 \texttt{SINGLE\_FACT\_LOOKUP}; 4 \texttt{LONG\_FORM\_SYNTHESIS}; all 10 have \texttt{hit\_at\_k=true}. \\
LongMemEval remaining task types & 12 & 156 & Single-session-user 2/70; single-session-assistant 3/56; single-session-preference 7/30; all 12 have \texttt{hit\_at\_k=true}. \\
MAB FC-MH & 229 & 400 & 214 \texttt{LONG\_FORM\_SYNTHESIS}; 15 \texttt{SINGLE\_FACT\_LOOKUP}. \\
MAB Movie-Rec & 179 & 200 & 179 \texttt{SINGLE\_FACT\_LOOKUP}. \\
MAB EventQA & 191 & 1500 & 188 \texttt{LONG\_FORM\_SYNTHESIS}. \\
MAB LongMemEval-S & 164 & 300 & 110 \texttt{SINGLE\_FACT\_LOOKUP}; 36 \texttt{TIME\_ANCHORED\_LOOKUP}. \\
MAB FC-SH & 72 & 400 & 65 \texttt{SINGLE\_FACT\_LOOKUP}. \\
MAB Detective-QA & 12 & 71 & 11 \texttt{SINGLE\_FACT\_LOOKUP}. \\
MAB InfBench-Sum & continuous & 100 & Mean summarization F1 32.77\%; min 0.00\%; max 91.30\%. \\
\bottomrule
\end{longtable}

These counts are not a human-labeled root-cause taxonomy. They are a reproducible sampling frame for manual failure inspection. The immediate use is to select examples from high-mass cells, inspect the retrieved chunks and predictions, and label each example as retrieval miss, evidence present but synthesis wrong, routing/procedure mismatch, abstention-normalization issue, or judge disagreement.

\subsection{Vector-store and synthesis-probe limitations}

The evaluation description uses pgvector-backed retrieval for \systemname{} and the production comparator and ChromaDB for Mem0. No paired ChromaDB-versus-pgvector control holds embedding model, \texttt{top\_k}, corpus scale, and scoring path constant. Vector-store choice therefore remains part of the configuration-level comparison.

The internal Sonnet probe is summarized as mean 4.22 seconds, median 3.96, p90 7.81, n = 10. Its inputs, per-call rows, script, timestamp, and model provenance are absent. More importantly, a constant offset cannot recover the missing per-question Stage-3 time for LoCoMo and LongMem. The main results therefore omit aligned latency claims for those scored paths.

\end{document}